# Hundred Drones Land in a Minute

Daiki Fujikura*, Kenjiro Tadakuma, Masahiro Watanabe, Yoshito Okada, Kazunori Ohno, and Satoshi Tadokoro

*Abstract*—Currently, drone research and development has received significant attention worldwide. Particularly, delivery services employ drones as it is a viable method to improve delivery efficiency by using a several unmanned drones. Research has been conducted to realize complete automation of drone control for such services. However, regarding the takeoff and landing port of the drones, conventional methods have focused on the landing operation of a single drone, and the continuous landing of multiple drones has not been realized. To address this issue, we propose a completely novel port system, "EAGLES Port," that allows several drones to continuously land and takeoff in a short time. Experiments verified that the landing time efficiency of the proposed port is ideally 7.5 times higher than that of conventional vertical landing systems. Moreover, the system can tolerate 270 mm of horizontal positional error, $\pm 30°$ of angular error in the drone's approach ($\pm 40°$ with the proposed gate mechanism), and up to 1.9 m/s of drone's approach speed. This technology significantly contributes to the scalability of drone usage. Therefore, it is critical for the development of a future drone port for the landing of automated drone swarms.

## I. Introduction

Recently, the field of drone research and development has come to the fore worldwide. Owing to their remarkable mobility, the use of drones increases in various industries, such as transportation, surveying, construction, and civil engineering, and enables labor saving, full automation, and improved services. Several major prominent companies are attempting to use unmanned aerial vehicles (UAVs) for commercial purposes, especially for package delivery. Prime The air service [1] provided by Amazon aims to realize package delivery to one's front door within 30 minutes from an order. Alibaba [2] and Wing [3], an affiliated company of Google, are also conducting test operations of their delivery services using UAVs.

Delivery services employing UAVs offer improved delivery efficiency by using a massive number of drones [4]. Package delivery using conventional transportation methods requires a high cost to improve delivery efficiency, which relies on equipment, such as labor, vehicles, vessels, and planes, that a business can afford. This often leads to scalability issues. Delivery to places with underdeveloped infrastructure also requires high cost for infrastructure

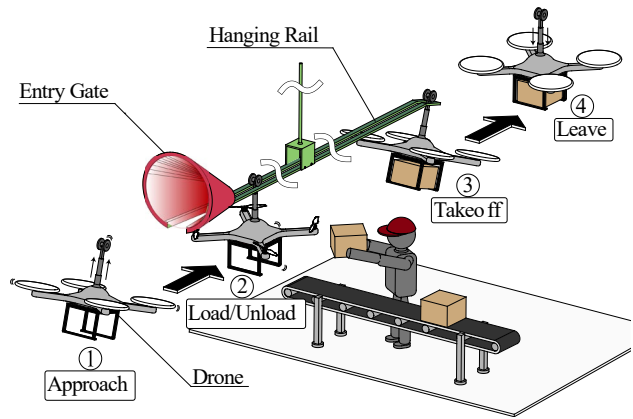

Fig. 1. Concept of proposed drone port "EAGLES Port"

establishment and human labors. In such cases, deliveries via drones is highly crucial.

Research is being conducted on various aspects to realize the full automation of drone control, namely the recognition of obstacles [5], creation of avoidance routes and surrounding maps [6][7], recognition of landing points [8], development of automatic landing system [9][10], simultaneous control of multiple drones [11]. However, regarding the drone takeoff and landing port, conventional methods aim for single-drone landing. A drone port that allows multiple drones to land continuously has not been realized, because of the long time required for landing and the complexity of the system.

In this study, we propose a novel port system, called "EAGLES Port", that enables a large number of drones to continuously land and takeoff in a short period of time. Its effectiveness is verified through experiments using a prototype. Under ideal operation, 100 drones per minute can land safely in the proposed port without complicated controls. Automation of the loading and unloading of packages is also realizable. This technology significantly contributes to the scalability of drone usage, and therefore, can be a critical technology for the development of future drone ports hosting automated drone swarms.

This paper describes the proposed hang-type drone port system, and its advantages and disadvantages against conventional landing methods are discussed in Section II. In Section III, the hardware design of the drone port, especially its entry gate, is described. The structure of the passive positioning mechanism that enhances the robustness of the landing is explained, and its principle is clarified through theoretical calculations. Additionally, an experiment is conducted using a prototype of the proposed entry gate mechanism, and the behavior of the gate are investigated, as presented in Section

*Daiki Fujikura, Kenjiro Tadakuma, Masahiro Watanabe, Yoshito Okada, Masanori Ohno, and Satoshi Tadokoro are with the Graduate school of Information Sciences, Tohoku University, Japan (email: fujikura.daiki@rm.is.tohoku.ac.jp). These authors contributed equally to this work.

TABLE I. COMPARISON OF LANDING MECHANISMS

| Mechanism | Continuous landing | Landing time | Approach speed before landing | Landing area | Attachment of additional structure | Landing stability | Precise control | Expandability for loading/unloading objects |
|---|---|---|---|---|---|---|---|---|
| Flat Plane | Requires special structure | At least 4.9 s [1] | Desired to be zero | Approximately 4 m$^2$ [4][12] | Basically none | Stable/ Difficult in the wind | Limited | Can be automated with a belt conveyor |
| Grippers [13][14] | Difficult | Depends on the landing speed | Depends on the landing speed | Special port | Requires. 0.1 kg–0.48 kg [13][14] | Depends on the environment | High | Difficult |
| Vertical wall [15][16] | Difficult | Approximately 3 s [2] | Require more than 0.4 m/s | Anywhere on smooth walls | Small adhesive pads | Difficult to land repeatedly | Not required | Difficult |
| Capture net [18][19] | Difficult | Takes time to control | Need to adjust the speed | Lands in the air | Basically none | Stable | Slightly | Difficult |
| Active pin array chamber | Difficult | 0.8 s [3] | Approximately 0.6 m/s [3] | Approximately 1 m$^2$ (twice as large as drone) [3] | Nets for hooking pins | Stable | Slightly | Possible |
| Hanging hook [21] | Difficult | Significantly short | Significantly high (installs shock absorber) | 2.4 m × 6 m (can be miniaturized) [22] | Small part to be hooked | Stable | Slightly | Requires an arm to drop the drone to the ground |
| Hanging rail (proposed) | Possible | 0.5 s | 1.9 m/s | 0.35 m × 0.32 m (gate size only) | Small part to slide in | Stable | Slightly ±0.14 m, ±30° | Can be automated with a belt conveyor/ Enough space at the bottom |

[1] Vertical landing from 1.5 m height. Experiment D.
[2] Calculated from the movie (https://www.youtube.com/watch?v=ySK_pvzatsk) [17].
[3] Calculated from the movie (https://www.youtube.com/watch?time_continue=176&=&v=7xYbhL4RY84) [20].
[4] Hoodman, HDLP

IV. In Section V, the results of the experiment, scope of future experiments, and their verification items are discussed. Finally, the conclusions and future deployment of the proposed drone port system are presented in Section VI.

## II. CONCEPT OF HANG-TYPE DRONE PORT SYSTEM

An overview of the proposed drone port system is illustrated in Fig. 1. The port comprises an entry gate that guides the drone to the entrance of the rail, a hang-type rigid rail, and a conveyor belt that transports the drone. The drone is equipped with a rod on top; this rod has a simple slippery tip structure to ensure that the drone can slide into the rail.

The drone enters the port and lands on the rail in the following procedure. First, the drone detects the entrance of the port using its sensors and inserts the tip structure of the attached rod into the wide entry gate. The tip structure is then guided by the positioning mechanism adapted to the gate to the position where the hanging rail is located. The mechanism constrains the position and the posture of the drone, enabling the compensation of approach position and approach angle errors. Finally, the drone ceases by stopping its propellers. The drone is then transported by the conveyors on a rail to perform various tasks, such as loading, unloading, and battery replacement.

The key features of the proposed system are the landing efficiency, necessary for multiple drone operation, and the high expandability and flexibility. Generally, conventional landing methods aim only for a single drone operation. For example, Flat Plane on Table I is the most adopted method for vertical landing toward a plane. It is the simplest landing method that only requires a flat ground. However, since the drone has to decrease the propeller speed to complete the landing task, this method is weak against disturbances like gusts. Therefore, the landing becomes unstable owing to the ground effect, takes a relatively long time to land, and requires a spacious port for safety.

The Gripper is a unique method that completes the landing task by grasping a fixed rod-shaped object. "Multistory parking" may be possible with this method since the drone does not have to land on the ground. However, the difficulty exists in the control of its position and posture upon landing, and the heaviness of the gripper itself decreases the maximum payload of the drone. Further, the gripper installed at the bottom of the drone makes package loading challenging, meaning that this method is not suitable for delivery services.

In the Vertical Wall, the drone lands on a vertical wall. The method enables drone landing and takeoff at special environments, similarly to the Gripper system. Its working environment is limited to places with a vertical wall where a drone can be attached to. Therefore, there is a risk that the drone may detach, and a long time to land and takeoff is required.

The Capture Net is a method to capture a drone in the air. Basically, the method is adopted in a situation where the drone in the air must be captured to get retrieved. Therefore, the method is inefficient and unsuitable for general landing use. The Active Pin Array Chamber is another method to capture a drone in the air, but unlike the Capture Net method, the drone is constrained and securely landed by the upper and lower rigid pins. However, the system is operated with a single drone.

The Hanging Hook method uses a hook attached to the upper part of the drone to complete landing. The drone lands by hooking to a wire placed above the drone. This method can be used with high-speed aircraft, like rigid wing UAVs, requires small space for port establishment, and leaves drones unaffected by ground effects. However, it is difficult to conduct consecutive landings with multiple drones with this method.

Contrarily, the proposed Hanging Rail method features the following advantages over conventional methods: i) it enables continuous landing of multiple drones in a short time without complex control, ii) it is a robust to disturbances such as gusts and stable upon landing, because the propeller speed can be maintained, iii) the port can be miniaturized as compared to existing ports and is more accessible to perform tasks such as loading/unloading of packages, maintenance, and battery exchange, as the port does not occupy the floor, iv) processes such as system operation, drone maintenance, and parking can be automated because the rail shape can be arbitrarily configured, and v) a roof can be installed for protection from rain.

### III. TAPERED GATE WITH SELF-ALIGNING ADAPTIVE MECHANISM

#### A. Position Compensation Mechanism

Compensation for the approach position and angle error of the drone is necessary to complete a quick landing. Conventional methods have taken several approaches to realize similar compensations.

Remote center compliance (RCC) is a positioning mechanism that is attached to the hand of a manipulator for a peg-in-hole assembly [23][24]. Horizontal position and angular errors are compensated by a rotation center virtually placed at the tip of the peg using diagonally placed links. The mechanism must be placed at the tip of the drone if it is to be used for the position angle compensation of the drone. This results in a largely decreased drone payload owing to the complexity of the structure. Therefore, implementation of RCC to delivery drones may be difficult.

Compensation of position and posture is vital for the rendezvous and docking operation of spacecrafts [25][26]. A total of 5 degrees of compensation is achieved, consisting of 2 degrees of planar position and 3 degrees of rotation. By the end of the operation, the position and posture are fully constrained. However, the operation time is of the order of minutes. For the drone error compensation, only a horizontal position and one axis of rotation need to be restricted, but the operation has to be quick. Thus, the method used for spacecrafts does not fulfill the requirements for drones, especially in terms of quickness.

#### B. Self-aligning Adaptive Mechanism

This study proposes a one degree of freedom (DOF) passive positioning mechanism with four-bar linkage system for position and orientation compensation. A schematic diagram of the proposed mechanism is shown in Fig. 2. Two curved links with the center of rotation arranged at the entry are connected to each other at the opposite end with a connecting member. This member is then hung by a slider located above, realizing a one DOF freedom.

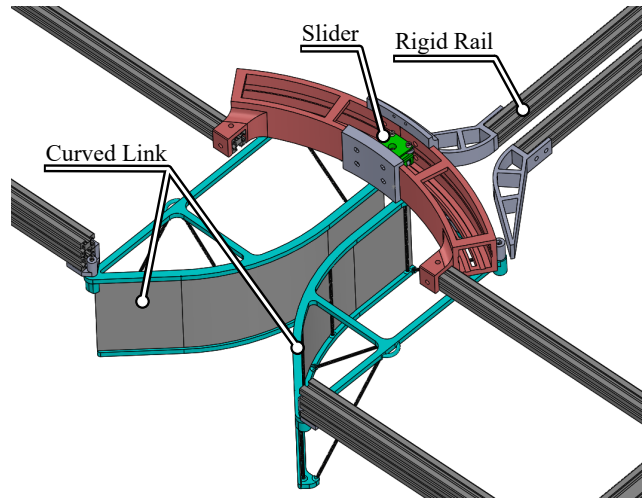

Fig. 2. Components of 1 DOF passive positioning mechanism shown with guide and rigid rails.

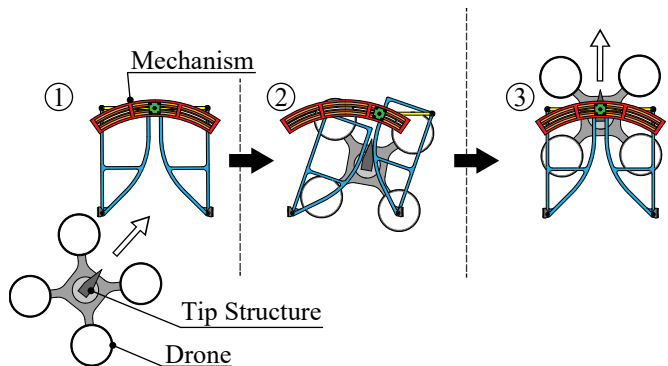

Fig. 3. Sequence of the estimated entry of the drone towards the proposed positioning mechanism

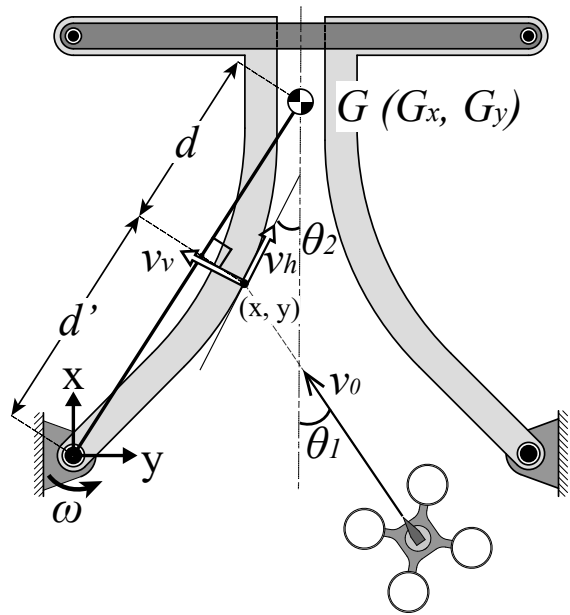

Fig. 4. The proposed mechanism is modelized with simple links and joints. The origin, axes, and other variables used to formulize the entry of a drone is defined.

Fig. 3 shows a sequence of the drone's approach toward the entry gate using the proposed mechanism. First, the drone with the approach position and angle error collides with the curved link of the mechanism. Then, the mechanism passively changes the taper opening angle of the curved link through the impact of the collision, making it easier for the drone to move into the gate. The width of the gate changes as the opening angle changes. To deal with this, the structure to be mounted on the drone was tapered so that the drone could enter the narrowed gate.

## C. Modeling

The entry gate comprising the proposed mechanism is modelized, as shown in Fig. 4. Subsequently, the entry of the drone is formulized. Frictional resistance is ignored in the model, and the drone is considered as a mass point. We assume that $\theta_1$ is the approach angle of the drone, $\theta_2$ is the opening angle of the link at the collision position, m is the mass of the drone, $v_0$ is the drone velocity before the collision, $I$ is the moment of inertia of the proposed mechanism, and $e$ is the coefficient of restitution upon collision. The origin is set to the rotation center of the link with which the drone collides; the y axis is directed toward another rotation center, and the x axis is directed toward the rail of the port. The properties of the collision are derived as follows:

$$\bar{f} = \frac{m(1+e)I}{I+md'} v_0 \sin(\theta_1 + \theta_2). \tag{1}$$

$$v_v = \left(1 - \frac{(1+e)I}{I+md'}\right) v_0 \sin(\theta_1 + \theta_2). \tag{2}$$

$$v_h = v_0 \cos(\theta_1 + \theta_2). \tag{3}$$

$$\omega = \frac{md'(1+e)}{I+md'} v_0 \sin(\theta_1 + \theta_2). \tag{4}$$

Here, $\bar{f}$ represents the impulse applied from the drone to the mechanism upon collision, $v_v$ is the drone velocity component after the collision that is perpendicular to the link at the collision position, $v_h$ is the same component but to the horizontal direction, and $\omega$ is the angular velocity of the mechanism. $d'$ is the distance from the rotation center to the point of intersection of the vertical line drawn from the collision position of the drone to the straight line connecting the rotation center and the center of gravity of the mechanism. The distance is obtained from a different formula depending on the relationship between the center of gravity and the collision position, as follows:

$$d' = d_G \pm d$$

$$= \begin{cases} \sqrt{G_x^2 + G_y^2} - \sqrt{(G_x^2 - x)^2 + (G_y - y)^2 - \frac{\left(\frac{G_y}{G_x}x - y\right)^2}{\left(\frac{G_y}{G_x}\right)^2 + 1}} \\ \sqrt{G_x^2 + G_y^2} + \sqrt{(G_x^2 - x)^2 + (G_y - y)^2 - \frac{\left(\frac{G_y}{G_x}x - y\right)^2}{\left(\frac{G_y}{G_x}\right)^2 + 1}} \end{cases} \tag{5}$$

Here, $d_G$ represents the distance from the rotation center to the center of gravity, and d represents the distance from the gravity center to the intersecting point of the vertical line from the collision position and the straight-line connecting the center of gravity and the rotation center of the mechanism. $G_x$ represents the x component of the gravity center, and $G_y$ represents the y component. The collision point is at x, y. The condition for using the top formula (5) is

$$G_x x + G_y y \leq G_x^2 + G_y^2, \tag{6}$$

and the condition for using the bottom formula is

$$G_x x + G_y y > G_x^2 + G_y^2. \tag{7}$$

The derived formulas suggest the entering condition of the drone to the entry gate with the following inequalities:

$$\theta_1 + \theta_2 < \frac{\pi}{2}. \tag{8}$$

$$md' > eI. \tag{9}$$

## IV. EXPERIMENTS

### A. Experimental Setup

An experiment is conducted to verify the performance of the drone port with the proposed mechanism adapted at the entry gate. First, the behavior of the drone upon gate entry is analyzed using video footage. Thereafter, two comparative experiments are conducted using a gate with the proposed mechanism and a fixed gate of the same shape. The success/failure rate of entry was analyzed to verify the advantages of the proposed mechanism, and the required landing time was measured to determine the influence of the proposed drone port.

The equipment and the drone equipped with a rod and tip structure used in the experiments are shown in Fig. 5 and Fig. 6, respectively. Their specifications are shown in Table II and Table III, respectively.

TABLE II. SPECIFICATIONS OF THE ENTRY GATE

| Entrance Width | 270 mm |
|---|---|
| End Width | 30 mm |
| Gate Depth | 280 mm |
| Height | 112 mm |
| Taper Angle | 45° |
| Mass of One Link | 120 g |

TABLE III. SPECIFICATIONS OF THE DRONE

| Model | DJI MAVIC AIR |
|---|---|
| Size | 168*184*64 mm (L*W*H) |
| Mass of the Drone | 430 g |
| Rod Length | 180 mm |
| Total Mass | 501.8 g |

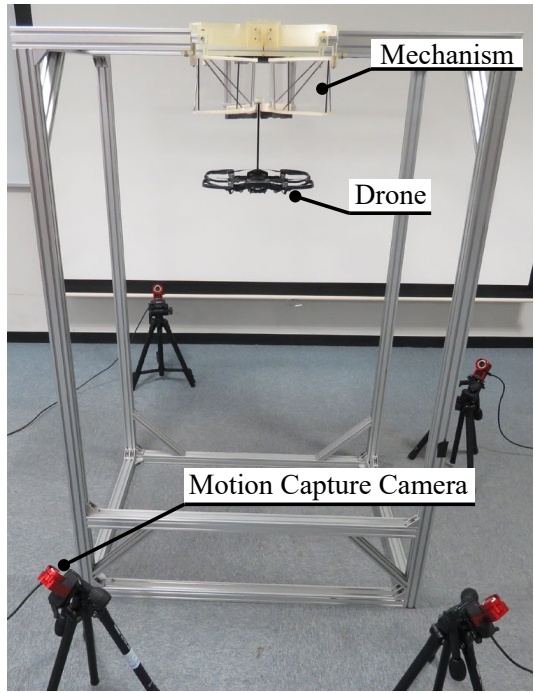

Fig. 5. Experimental setup

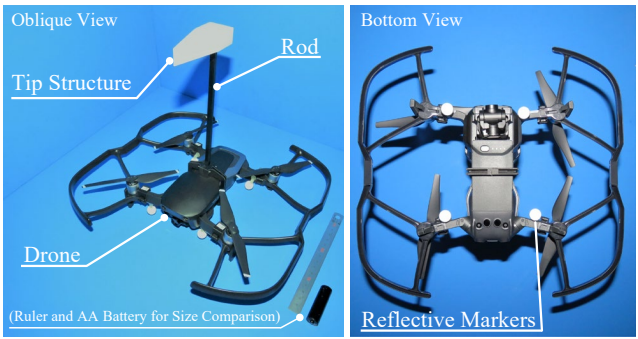

Fig. 6. Drone configuration and the position of its reflective markers

### B. Movie Analysis of the Motion of the Proposed Gate upon Drone Entry

In this section, the behavior of the drone upon entry is analyzed. The drone was given only the maximum advancing input, which was maintained until the drone reached the rigid rail or bounced back against the link of the gate. The state of entry of the drone is shown in Fig. 7. The gate, which was originally positioned at its initial location, was rotated by the collision with the drone, decreasing its taper opening angle. Then, the drone entered into the gate by pushing and widening the narrowed gate width, was guided toward the center of the gate, and entered the rigid rail installed beyond the gate. This behavior was in accordance with the theoretical modeling of the mechanism was proposed; therefore, this mechanism can be adopted as the entry gate of the proposed drone port.

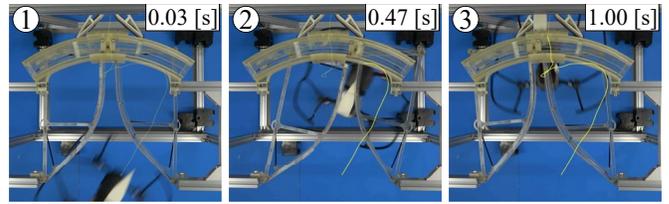

Fig. 7. Sequence of entry in the experimental procedure shown with a yellow spline that represents the estimated trajectory of the head of the drone

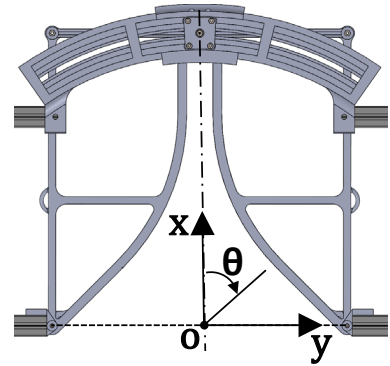

Fig. 8. Definition of the origin and axes used to analyze the result of the experiment C

### C. Comparison of Successful/Failed Entry Ratio

An experiment was conducted to compare the performance difference between the entry gate with the proposed mechanism and a simple gate using fixed a tapered rail of the same shape. The drone was made to enter randomly into the right side link/rail of each gate for 100 times. For each trial, the result (either a successful or failed entry), approach angle, and approach speed of the drone were recorded using video and motion capture. A successful entry was defined for situations when the drone passed through the entry gate and reached the rigid rail or the guide rail. Any other situations, such as when the drone bounced back against the gate or could not reach the rigid rail, was characterized as a failed entry. Upon its approach toward the gate, the drone was given only the maximum advancing input and was maintained until the defined success and failure requirements were met. Additionally, the origin, x axis, and y axis were defined as shown in Fig. 8.

The experimental results of each entry gate are shown in Fig. 9 as a scatter diagram grouped by success and failure with the entering velocity on the horizontal axis and the entering angle on the vertical axis. By comparing the two graphs, all the trials with entering angle of over 40° failed to enter. A difference in the successful entry ratio was seen for an entering angle of 35–40°. For the entry gate with the proposed mechanism, we observed two cases of failure, where the drone popped out from the upper side of the entry gate and got stuck on the frame that fixes the gate at an entering velocity around 1.8 m/s and an entering angle of 20–25°. Such failure can be avoided by adding a roof to the proposed mechanism.

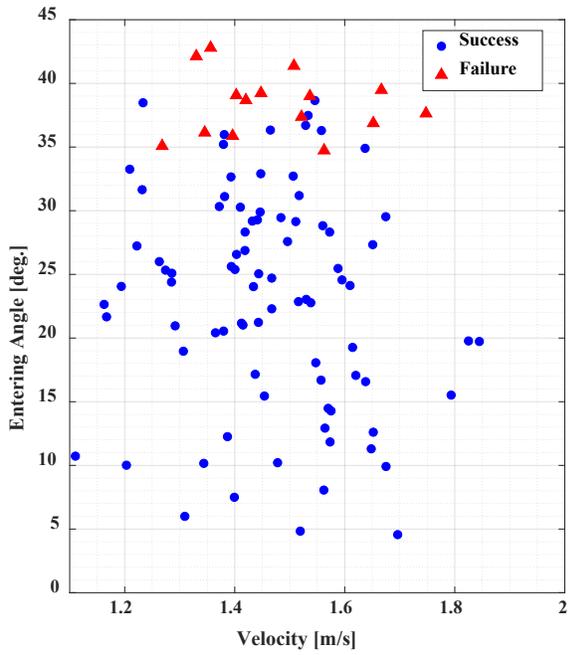 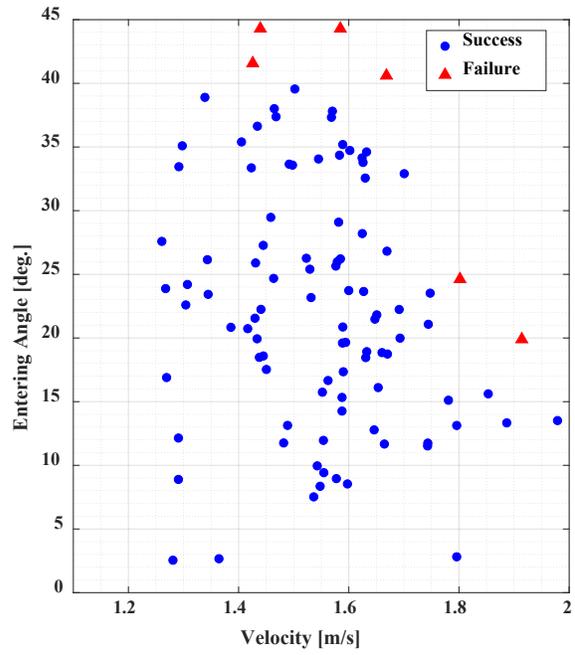

(a) Fixed tapered entry gate  (b) Entry gate with positioning mechanism

Fig. 9. Scatter graphs of the experimental results showing the successful/failed entry of each trial

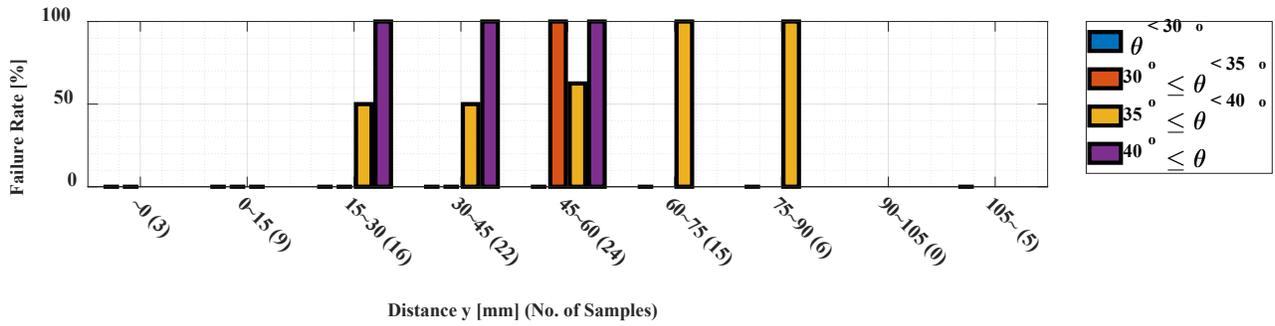

(a) Fixed tapered entry gate

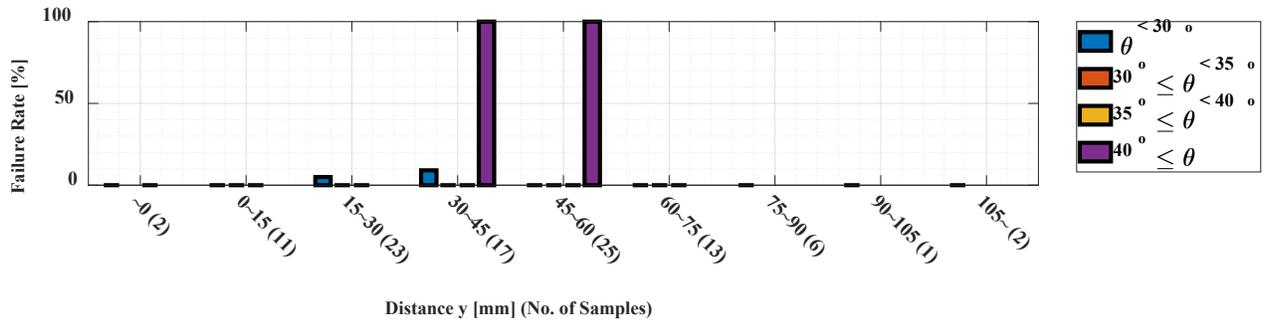

(b) Entry gate with positioning mechanism

Fig. 10 Result of the experiment expressed as bar graphs which represents the failure rate grouped by an entering angle and a collision position

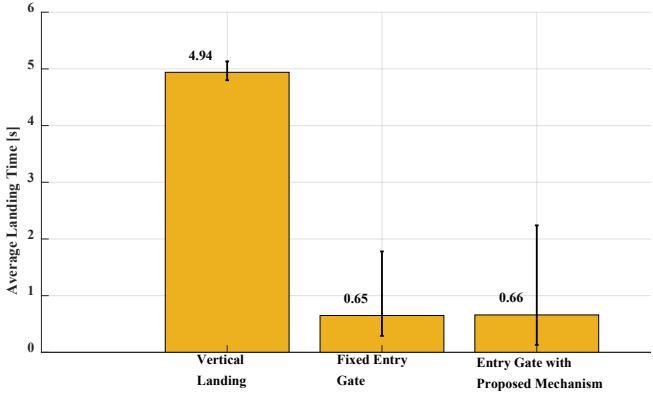

Fig. 11 Lading time compared as bar graphs

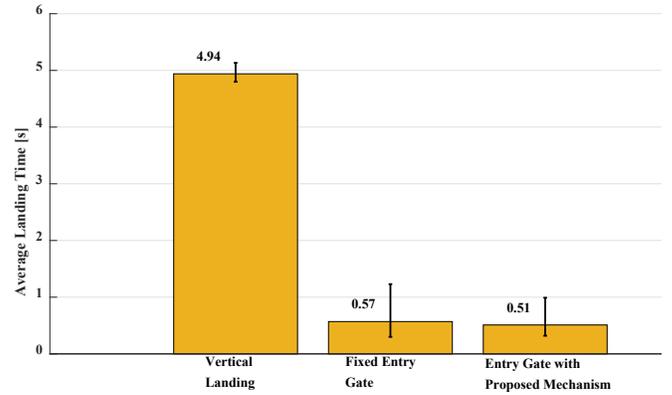

Fig. 12 Revised landing time compared as bar graphs

Next, to further analyze the difference seen in the previous graphs, the same results are shown in Fig. 10 as a bar graph grouped by the range of the entering angle with the failure rate on the vertical axis and the y component of the center-of-gravity location of the drone upon collision grouped in the range of 15 mm on the horizontal axis. The difference in the successful entry ratio that was seen in the previous scatter graphs can now be seen as a failure ratio on the bar graphs. The observed difference can be seen in the graph of the fixed entry gate at the collision position range of 15 to 90 mm, which is more than 60% of the total width of one link/rail. In this whole range, more than or equal to 50% of the failure rate is observed for an entering angle of 35–40°, meaning that the difference is universally observed owing to the mechanical difference of the entry gate.

Therefore, it is confirmed that implementation of the proposed mechanism to the entry gate can tolerate an entering angle error of approximately 5° larger than that of the fixed tapered gate.

### D. Comparison of Landing Time

The time required for landing was measured and compared for the vertical landing in the proposed drone port with a fixed entry gate and for the drone port with the proposed passive positioning mechanism. For the vertical landing, the tip structure attached to the drone by a rod was positioned by an aluminum frame at the same height as the entry gate of the drone port. The landing time was defined as the time from the moment when the drone's tip structure left the frame to the moment when one of its legs touched the ground. For the proposed drone port, the landing time was defined as the time from the moment when the drone passed the point x=-100 mm to the moment when the drone reached the point x=410 mm, where the rigid rail first appears. Sample data was taken from all successful entry trials in which the motion capture data was fully obtained up to 410 mm from the previous experiment.

The results of the experiment are shown in Fig. 11. There is almost no difference in the landing time between the fixed entry gate and the gate with the proposed mechanism, confirming that there was little to no effect on the landing time with or without the mechanism. Moreover, the time required for landing in the proposed drone port was approximately 4.3 s shorter. A simple calculation shows that the landing efficiency is 7.5 times higher than that of vertical landing.

## V. Discussion

Further consideration is needed to determine the maximum entering velocity to the entry gate using the proposed mechanism. An experiment with higher drone velocity is necessary. The experiment using the entry gate with the proposed mechanism recorded successful trials with an entering velocity of up to 1.9 m/s and 1.8 m/s for the fixed entry gate. However, these successful trials were only confirmed when the entering angle was less than 20°. Trials at the same velocity range but with bigger entering angles should be tested to verify whether successful entry with such entering velocity is possible at a bigger angle range.

A limit on the acceptable approach speed and approach angle should be set when assuming the actual operation in the drone port. In real operations, it is uncommon for a drone to approach with an angle error of 30° or more. For example, the limit of the maximum entering velocity and entering angle for the drone port with the proposed mechanism could be 1.6 m/s and 30°, respectively. In the experiments, trials within this range had a successful entry ratio of 100%. Furthermore, the average landing time within this range is revised in Fig. 12, suggesting that the continuous landing efficiency can be even higher. Under this limitation, the proposed drone port system can ideally realize more than 100 drone landings per minute.

It is conceivable that the center positioning performance of the proposed mechanism changes depending on the length of the straight part of the curved link. In the experiments, the rod of the drone came in contact with the guide rail attached to the rigid rail at most trials. This can be solved by further increasing the length of the straight line. Moreover, it is anticipated that the length of the straight part and the performance of the positioning accuracy are related via mathematical formulas. Therefore, the optimal length of the straight part can be derived as a theoretical value; this theoretical value can later be compared with the experimental results.

## VI. Conclusion

This paper proposes a novel drone port system, "EAGLES Port", that can perform successive and secure drone landings. The landing time efficiency of the proposed port is ideally 7.5 times higher than that of conventional vertical landing systems. Moreover, a 1 DOF passive positioning mechanism is proposed and adopted for the entry gate of the port. Using the proposed mechanism, the system can tolerate a 5° greater entering angle error, as compared to the fixed tapered entry gate.

In the future, we intend to analyze the collision of the drone and the entry gate and determine the entry condition for a drone based on the established model formulas. Additionally, an experiment will be performed at a higher speed to determine the drone's limit approach velocity. Simultaneously, the optimal straight-line length of the entry gate will be elucidated by mathematical formulas and experiments to realize higher positioning accuracy. The port structure will be redesigned to withstand higher drone approach speeds.

A new mechanism that can also compensate for the position error in the height direction will be devised in the future by expanding the mechanism proposed in this paper, to improve the stability on landing. Furthermore, the entire system of the port will be built and automated, including the tasks of package loading/unloading.